# BrainNet-MoE: Brain-Inspired Mixture-of-Experts Learning for Neurological Disease Identification

Jing Zhang[1] Xiaowei Yu[1] Tong Chen[1] Chao Cao[1] Mingheng Chen[1] Yan Zhuang[1]
Yanjun Lyu[1] Lu Zhang[2] Li Su[3,4] Tianming Liu[5] Dajiang Zhu[1]

[1]Computer Science and Engineering, The University of Texas at Arlington, Arlington, TX, USA
[2]Department of Computer Science, Indiana University Indianapolis, IN, USA
[3]Department of Psychiatry, School of Clinical Medicine, University of Cambridge, UK
[4]Neuroscience Institute, School of Medicine and Population Health, University of Sheffield, UK
[5]School of Computing, The University of Georgia, Athens, GA,
`jxz7537@mavs.uta.edu`, `xxy1302@mavs.uta.edu`,
`txc5603@mavs.uta.edu`, `cxc0366@mavs.uta.edu`,
`mxc2442@mavs.uta.edu`, `yxz8653@mavs.uta.edu`,
`yxl9168@mavs.uta.edu`, `lz50@iu.edu`, `l.su@sheffield.ac.uk`,
`tliu@uga.edu`, `dajiang.zhu@uta.edu`

**Abstract.** The Lewy body dementia (LBD) is the second most common neurodegenerative dementia after Alzheimer's disease (AD). Early differentiation between AD and LBD is crucial because they require different treatment approaches, but this is challenging due to significant clinical overlap, heterogeneity, complex pathogenesis, and the rarity of LBD. While recent advances in artificial intelligence (AI) demonstrate powerful learning capabilities and offer new hope for accurate diagnosis, existing methods primary focus on designing "neural-level networks". Our work represents a pioneering effort in modeling system-level artificial neural network called **BrainNet-MoE** for brain modeling and diagnosing. Inspired by the brain's hierarchical organization of bottom-up sensory integration and top-down control, we design a set of disease-specific expert groups to process brain sub-network under different condition, A disease gate mechanism guides the specialization of expert groups, while a transformer layer enables communication between all sub-networks, generating a comprehensive whole-brain representation for downstream disease classification. Experimental results show superior classification accuracy with interpretable insights into how brain sub-networks contribute to different neurodegenerative conditions.

## 1 Introduction

Characterized by a progressive decline in cognitive function that significantly interferes with daily life, dementia presents major clinical and socioeconomic challenges. Accurate differential diagnosis is crucial for providing targeted treatments and slowing symptom progression, as different types of dementia vary in prognosis and response to medication, however, differentiating between dementia types is challenging due to overlapping symptoms [1-2]. Despite being the second most common neurodegenerative dementia after Alzheimer's disease (AD), Lewy body dementia (LBD) remains underdiagnosed because its symptoms often resemble those of AD and other dementia types [3-6]. Distinguishing LBD and AD from Normal Controls (NC) is essential not only for understanding how different pathologies affect brain structure but also for enabling early intervention and targeted drug treatment. Previous efforts on LBD have primarily relied on group-wise statistical methods to analyze structural connectivity networks [6-7], largely due to the rarity of LBD cases and the difficulty in acquiring sufficient data for deep learning approaches. While study [8] used EEG signals to train a Long Short-Term Memory (LSTM) model for multi-disease classification but EEG data suffers from low spatial resolution and high temporal variability, limiting its reliability. Ni et al. [9] trained deep learning (DL) models to distinguish AD from LBD using Tc-99m-ECD SPECT images, but such nuclear imaging techniques are costly and have limited accessibility. Wang et al. [10] and Nemoto et al. [11] proposed DL-based framework to detect neuroimaging signatures linked to different pathologies, but their reliance of T1-weighted brain scans faces challenges of low disease-related information and high level of noise.

The human brain, with its remarkable ability to process, integrate, and adapt to information, has long served as a source of inspiration for artificial intelligence (AI). The evolution of artificial neural networks (ANNs) spans from early multilayer perceptrons (MLPs), which inspired by the brain's feedforward signal processing [12], to the recent Transformer architecture, whose key attention mechanism draws inspiration from human visual system [12-14]. Given their powerful representation capabilities, ANNs have been widely adopted in brain science, helping us understand human brain mechanism and enabling various real-world healthcare applications like early brain disease detection [15-17], brain age prediction [18-19], brain damage assessment [20]. Despite these advances, several challenges remain. Bridging insights from biological neural networks to inspire ANN architecture design remains an open challenge. Furthermore, effectively leveraging ANNs for brain modeling with limited neuroimaging data, particularly for rare disease like LBD diagnosis, remains largely underexplored.

While traditional ANNs focused on designing "neural-level networks" with primarily linear organization, exemplified by ResNet's sequential convolutional neural networks (CNNs) [21] and Transformer's stacked multiple layers [22]. In our work, we pose a different question: *Can we design a system-level architecture that mimics the brain's fundamental principle where different regions specialize in distinct tasks while working in an integrated manner, and further helping understand complex brain patterns?* This question leads us to develop **BrainNet-MoE**, where multiple specialized experts collaborate to encode brain networks. Specifically, experts within and across different disease groups are initially identical, with each group collectively encoding a brain sub-network. These experts are integrated through a routing mechanism that mimics the human brain's selective activation patterns, generating latent space representation. Acting as a soft activator, a disease gate then assigns weights to these sub-net-works, functioning like the prefrontal cortex and thalamus in regulating resource allocation across brain regions [23]. This gate determines the association strength between brain sub-network and specific diseases, generating a disease-weight vector. When multiplied with the different expert groups' sub-network representations, this produces disease-informed representations. After processing all brain sub-networks of a subject, considering the collaborative and reciprocal nature of brain connectivity, a transformer layer is employed to enable communication between all sub-networks, generating a whole-brain representation for downstream disease classification.

An interesting brain-like property of our BrainNet-MoE is that during training, expert groups gradually specialize into disease-specific roles, mirroring how brain regions differentiate during development and progressively optimize for distinct cognitive functions. Similarly, the disease gate is designed to assess global brain information, it continuously improves its understanding of brain networking throughout training, becoming increasingly adept at allocating expert groups effectively. This global perception resembles how the brain dynamically adjusts the activation patterns based on cognitive tasks. To validate our BrainNet-MoE, we utilize brain Structural Connectivity (SC) as a whole-network because SC provides the anatomical foundation and serves as a valuable indicator of neurological disease progression [24-26]. One brain region connected to all the other brain regions represent a brain sub-network. We evaluated BrainNet-MoE on the challenging downstream task of distinguishing between AD, LBD, and NC, the results show superior compared to baseline.

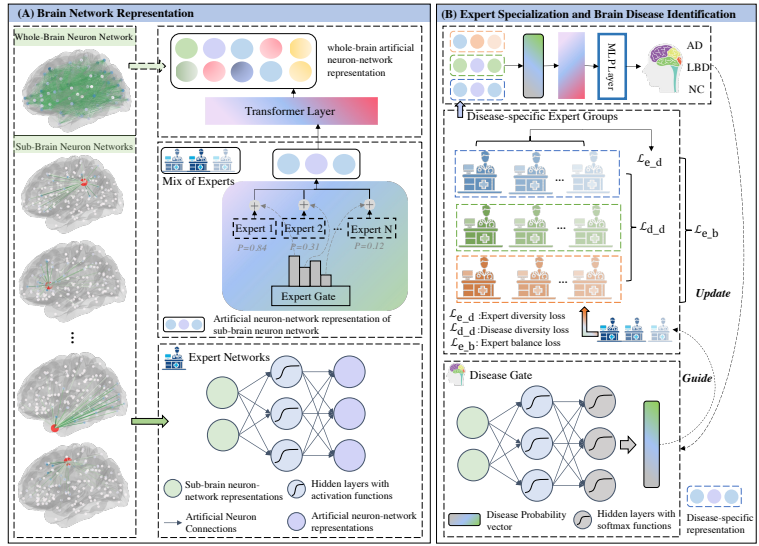

**Fig. 1.** Overview of the proposed framework, which consists of two main components: (1) Brain network representation, using a mixture of expert networks to model brain connectomics, and (2) Expert specialization and brain disease identification, where general experts gradually evolve into disease-specific expert groups during training, facilitating downstream diagnostic tasks.

## 2   Methodology

As illustrated in Fig. 1, our proposed BrainNet-MoE includes two main parts: (1) Brain Network Representation and (2) Expert Specialization and Brain Disease Identification. The first module leverages a mixture of expert networks to effectively capture brain connectome. Meanwhile, the second module enables general experts to progressively evolve into disease-specialized expert groups, facilitating downstream diagnosis tasks.

### 2.1   Brain Network Representation

The brain whole-network is defined as a SC network that derived from white matter fiber tracts in DTI and quantified by fiber tract count, denoted as $SC \in R^{N \times N}$, where $N$ is the number of brain regions. Consequently, the brain sub-network is defined as one brain region connected to all the other brain regions, with fiber counts representing connection strength, serving brain structural connectivity properties. To address highly skewed distribution of SC, we applied a logarithmic transformation followed by normalization as $SC = \frac{log_2(SC+1) - SC_\mu}{SC_\sigma}$ to standardize the data distribution, where $SC_\mu$ and $SC_\sigma$ are the mean and the standard deviation of $SC$, respectively.

Next, for a given brain whole-network, each expert network within a mixture-of-expert (MoE) group sequentially processes individual brain sub-networks $X$. Specifically, the expert network $f$ is a MLP neuron network with activation functions, enabling a nonlinear representation of the brain network in latent space. The MoE layer activates all expert networks $\{f_1, f_2, \ldots, f_N\}$ in each iteration, mimicking the human brain's selective activation patterns that dynamically allocating computational resources to different experts based on the input and generating brain sub-network representation $V_{sub-Brain}$. This process can be formulated as:

$$V_{sub-Brain} = F_{MoE}(x; \Theta, \{W_i\}_{i=1}^N) = \sum_{i=1}^N \mathcal{G}(x; \Theta)_i f_i(x; W_i) \quad (1)$$

$$\mathcal{G}(x; \Theta)_i = softmax(g(x; \Theta))_i = \frac{exp(g(x;\Theta)_i)}{\sum_{j=1}^N exp(g(x;\Theta)_j)} \quad (2)$$

Where $g(x; \Theta)$ represents the gating value prior to the softmax operation, determining the contribution of each expert.

Once we obtain all the brain sub-network representations $V_{sub-Brain}$, a key step is integrating them to formulate a whole-representation. Considering that human brain regions operate collaboratively and neurological disease typically affect across multiple regions, we implement a transformer layer to model inter-network communications between all sub-networks and generate a comprehensive brain representation as $V_{Brain} = softmax(\frac{QK^T}{\sqrt{d}} V)$, where $Q, K, V$ are matrices derived from $V_{sub-Brain}$. With the transformer's ability in capturing sequential relationships between all brain sub-networks, it not only incorporates information from individual sub-networks but also model their intricate interactions.

## 2.2 Expert Specialization and Brain Disease Identification

**Expert Specialization.** One key mechanism in our design is simulating the functional differentiation of brain regions. By design, expert groups gradually specialize into disease-specific roles, mirroring how brain regions differentiate during development and progressively optimize for distinct cognitive functions. To achieve this, we introduce a ***disease gate***, which assess global brain information and generating a disease weight vector during each iteration. This vector is then multiplied with $V_{sub-Brain}$, determining the association strength between each sub-network and different diseases to produce a disease-informed brain representation $V_{disease}$:

$$V_{hidden} = \sigma\,(W_1 V_{sub-Brain} + b_1)\,;\quad V_{disease} = \text{softmax}\,(V_{sub-Brain} + b_2)\,)\quad (3)$$

Where $W$ and $b$ are weight matrix and bias in different feedforward layers.

This global perception mechanism is inspired by how human brain processes multiple sensory modalities in real-world environments. For example, the prefrontal cortex plays a crucial role in adapting to different situational contexts, enabling cognitive control and decision-making based on real-time perception [27]. Additionally, the thalamus processes and routes sensory information through specific nuclei to associated cortical areas [28]. These biological neural mechanisms are essential for regulating resource allocation across brain regions, and our ***disease gate*** draws inspiration from them. The softmax function creates a probability distribution reflecting the brain's collaborative nature, where regions work together with varying involvement rather than in isolation, even during specialized tasks. This design creates a synergistic feedback loop: as experts become more specialized, the disease gate refines its understanding of brain representations, which in turn enables more precise expert allocation. This mutual enhancement mirrors the brain's hierarchical organization, where bottom-up sensory integration and top-down control work in concert to enhance cognitive processing [29].

To further encourage meaningful expert specialization, we design a comprehensive loss function with multiple complementary objectives. First, we introduce an ***expert diversity loss***, that encourages experts to develop distinct specializations within the same disease group. Specifically, we implement an entropy constraint to explicitly penalize uniform distributions:

$$\mathcal{L}_{expert\_diversity} = \sum_{i=1}^{k}(1 - \text{std}(\mathcal{G}_i) + \lambda H(\mathcal{G}_i)) \quad (4)$$

Where $\mathcal{G}_i \in R^{B \times N}$ represents the gating weights for the $i$-th disease across batch size $B$ and sub-network $N$, $H(\mathcal{G}_i) = -\sum_j \mathcal{G}_{ij} log \mathcal{G}_{ij}$ represents the Shannon entropy, which discourages uniform distributions over experts and encourages confident selections.

Second, we introduce a ***disease diversity loss*** by maximizing the maximize inter-disease separation while maintaining intra-disease consistency, given the $V_{disease}$:

$$\mathcal{L}_{disease\_diversity} = \sum_{i \neq j} \text{sim}(V_{disease}^i, V_{disease}^j) - \text{sim}(V_{disease}^i, \overline{V}_{disease}^i) \quad (5)$$

Lastly, we introduce a ***balanced usage loss*** to ensure all experts remain active and contribute meaningfully by penalizing deviation from uniform expert utilization:

$$\mathcal{L}_{expert\_balance} = \mathcal{W}_1(p_{emp}, p_{uniform}) \tag{6}$$

Where $p_{emp}$ represents the empirical distribution of expert activations, $p_{uniform}$ is a uniform distribution ensuring equal utilization, and $\mathcal{W}_1$ is the Wasserstein-1 distance measuring the transport cost. This multi-objective optimization leads to a more robust and interpretable model where experts naturally differentiate into disease-specific roles while maintaining efficient resource allocation.

**Disease Identification.** After obtaining all disease-informed sub-brain representation $\{V_{d1}, V_{d2}, \ldots, V_{dN}\}$, the transformer layer integrates them to a whole-brain disease-informed representation $V_{Brain\_d}$. This representation serves as the input for the final classification layer, implemented as a MLP and cross-entropy loss $\mathcal{L}_{cls}$. The overall training objective is:

$$\mathcal{L} = \mathcal{L}_{cls} + \alpha \mathcal{L}_{expert\_diversity} + \beta \mathcal{L}_{disease\_diversity} + \gamma \mathcal{L}_{expert\_balance} \tag{7}$$

where $\alpha, \beta, \gamma$ are trainable hyperparameters to balance the contribution.

## 3  Experiments and results

### 3.1  Data Preprocessing and Experimental Design

In this study, we adopt the in-house dataset from (Anonymized) with 166 subjects (CN: 23, LBD: 77, AD: 66). Standard imaging preprocessing was performed as described in [30], including eddy current correction with FSL and fiber tracking using DSI Studio. Each T1-weighted image was aligned to its corresponding DTI space via FSL FLIRT and segmented with FreeSurfer. The cortex was then parcellated into 148 regions using the Destrieux Atlas and individual SC were built based on fiber counts.

The model consists of several components. Specifically, we designed three disease-specific expert groups, with each group containing three experts. Each expert network is a two-layer MLP with a hidden dimension of 256 and uses GELU activation. The disease gate is also a two-layer MLP, with an output dimension of 3 and a softmax activation function. The transformer module has two layers with a hidden dimension of 128. Training was conducted on a single NVIDIA A6000 GPU for 32 epochs with a batch size of 64, using the AdamW optimizer with a learning rate of 1e-4. To ensure robust generalization, we used 80% of the data for training and 20% for testing.

### 3.2  Result

**Identifying Critical Disease-Related Brain Sub-networks.** To provides interpretability and explore how specific brain sub-networks are affected by different neurodegenerative diseases, we compute a relevance score by combining $V_{disease}$, which quantifies the contribute to special disease of each sub-network, and $g(x;\Theta)$, which reflects the activation level of sub-networks. The results are presented as Fig. 2.

First, in the CN *vs* AD panel, the brain sub network *rh_G_cingul-Post-dorsal*, which belongs to Posterior Cingulate Cortex, is involved in various cognitive tasks and has been identified as an early-affected cortical region in AD, exhibiting structural atrophy, functional decline, and metabolic reduction [31-34]. Additionally, the analysis identified notable AD-related changes in the anterior Mid-Cingulate Cortex and the left Inferior Parietal Gyrus (IPG), both of which are strongly associated with cognitive control [35-36]. Next, in the CN *vs* LBD panel, the *lh_S_oc-temp_lat*, belongs to Lateral Occipito-Temporal regions in LBD has shown weakened functional connectivity and potentially leading to visual hallucinations [37-38]. The anterior occipital sulcus, a key landmark between the occipital and temporal lobes, is crucial for visual processing and integration, and its disrupted connectivity may impair perception, object recognition, and spatial awareness, contributing to hallucinations in LBD [39]. Notably, these findings also serve as a discriminative between AD and LBD, as shown in the panel of AD *vs* LBD. Furthermore, we find discriminative sub-networks distinguishing CN *vs* AD *vs* LBD in the last panel. All disease relevance scores are also displayed.

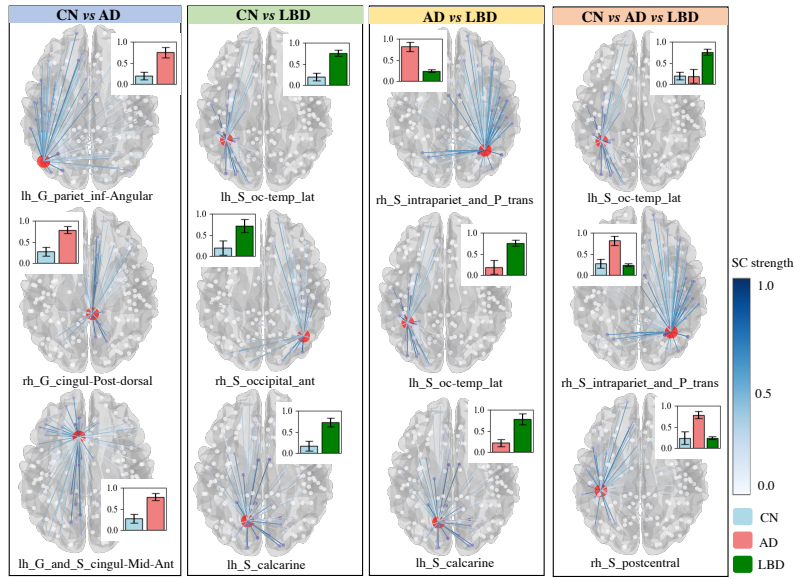

**Fig. 2.** Visualization of the Top-3 discriminative brain sub-networks across different diagnostic comparisons: CN *vs* AD, CN *vs* LBD, AD *vs* LBD, and CN *vs* AD *vs* LBD. Each sub-network is represented by one brain region (highlighted in red) connected to all the other brain regions (shown as bubbles), with structural connectivity strength indicated by color intensity. Inset bar plots show the discriminative scores assigned by expert groups.

**Evaluate Effectiveness on Downstream Diagnosis Task.** We compare *BrainNet-MoE* with four prominent conventional baseline models (KNN, Decision Tree, and XG Boost) and two transformer-based models: Transformer-Small (a 4-layer model with an embedding size of 256) and Transformer-Large (an embedding size of 512). Additionally, we evaluate two variations of ResNet (ResNet34 and ResNet50). As shown in Table 1. Overall, deep learning (DL)-based methods outperform conventional methods, and our proposed model surpasses all baselines. This work represents a pioneering effort in leveraging artificial neural networks (ANNs) for brain modeling and the diagnosis of rare diseases such as Lewy body dementia (LBD).

**Ablation Study.** First, we evaluated the impact of the number of experts in each disease-specific group. As shown in Table 2 (A), both increasing and decreasing the number of experts lead to a performance drop. This may be because too few experts limit pattern learning, while too many introduce redundancy or overfitting. Next, we assessed the individual contribution of each loss function. Table 2 (B) demonstrates that each loss component contributes meaningfully. Removing any individual loss term leads to decreased performance, with the most significant drop observed when all three specialized losses are removed. This validates our design of using complementary loss terms to guide expert specialization while maintaining balanced utilization.

. **Table 1.** Comparison of identification performance with other methods. ACC: Accuracy; SEN: Sensitivity; SPE: Specificity; PRE: Precision; F1: F1-score

| Models | ACC% | SEN% | SPE% | PRE% | F1% |
|---|---|---|---|---|---|
| SVM | 58.62 | 49.44 | 74.88 | 44.30 | 46.57 |
| Decision Tree | 48.28 | 43.61 | 73.29 | 48.08 | 44.76 |
| XG Boost | 62.07 | 53.61 | 79.50 | 42.39 | 46.67 |
| Transformer-small | 71.32 | 51.39 | 74.13 | 68.57 | 59.32 |
| Transformer-large | 67.82 | 33.33 | 66.67 | 17.24 | 22.73 |
| ResNet-34 | 70.11 | 54.44 | 78.05 | 52.72 | 48.19 |
| Resnet-50 | 72.41 | 43.61 | 72.22 | 43.67 | 40.0 |
| **BrainNet-MoE** | **82.76** | **88.89** | **91.38** | **82.69** | **80.07** |

**Table 2.** Quantitative results of ablation study. Part (A) evaluates the impact of number of experts in each disease-specific group. Part (B) examines each loss function. $\mathcal{L}_{e\_d}$, $\mathcal{L}_{d\_d}$, $\mathcal{L}_{e\_b}$ means expert diversity loss, disease diversity loss, balanced usage loss, respectively.

| Ablation Strategy | ACC% | SEN% | SPE% | PRE% | F1% |
|---|---|---|---|---|---|
| *(A) Number of experts ablation* | | | | | |
| 2 | 65.52 | 58.61 | 59.46 | 59.89 | 65.52 |
| 4 | 70.12 | 68.33 | 70.37 | 71.67 | 68.89 |
| *(B) Loss function ablation* | | | | | |
| w/o $\mathcal{L}_{e\_d}$ | 79.31 | 80.17 | 86.51 | 82.45 | 78.46 |
| w/o $\mathcal{L}_{d\_d}$, | 75.86 | 76.22 | 84.92 | 79.49 | 73.63 |
| w/o $\mathcal{L}_{e\_b}$ | 80.17 | 76.17 | 85.11 | 80.85 | 78.0 |
| w/o $\mathcal{L}_{e\_d}$, $\mathcal{L}_{d\_d}$, $\mathcal{L}_{e\_b}$ | 72.41 | 71.59 | 81.75 | 78.88 | 68.21 |
| **Ours** | **82.76** | **88.89** | **91.38** | **82.69** | **80.07** |

## 4      Conclusion

Our proposed **BrainNet-MoE** offers three key advantages: (1) Inspired by brain organization principles, it effectively captures complex brain patterns and demonstrates superior diagnostic performance for rare diseases; (2) it provides interpretable disease-specific brain patterns by revealing how different brain sub-networks contribute to various neurodegenerative conditions; and (3) it offers potential for generalization by incorporating new neurological disease patterns through additional disease-specific expert groups, similar to how the human brain adapts to new knowledge. In future, we will further explore complex brain representations from functional connectome.